%% file: main.tex
\documentclass[letterpaper]{article}
\usepackage{natbib,alifeconf}
\usepackage{alifeconf}
\usepackage{stfloats}
\usepackage[hyphens]{url}  
\usepackage{graphicx} 

\usepackage{natbib}  
\usepackage{caption} 
\usepackage{algorithm}
\usepackage{algpseudocode}

\algnewcommand\algorithmicforeach{\textbf{for each}}
\algdef{S}[FOR]{ForEach}[1]{\algorithmicforeach\ #1\ \algorithmicdo}

\usepackage{amsmath}

\newcommand{\mi}{\hat{\mathcal{I}}}

\newcommand{\pos}{\boldsymbol{p}}
\usepackage{romannum}
\usepackage{amssymb}
\usepackage{cancel}

\usepackage{booktabs}
\usepackage[table]{xcolor}
\definecolor{grey}{rgb}{0.8,0.8,0.8}
\definecolor{lgrey}{rgb}{0.9,0.9,0.9}
\definecolor{mypink}{rgb}{0.858, 0.188, 0.478}

\usepackage[labelformat=simple]{subcaption}

\usepackage{amsmath} 
\usepackage[acronym, toc, shortcuts, nowarn]{glossaries}

\title{Reliably Re-Acting to Partner's Actions with the Social Intrinsic Motivation of Transfer Empowerment}

\author{
\bf Tessa van der Heiden \\ BMW Group \\ tessa.heiden@bmw.de
\And Herke van Hoof \\ University of Amsterdam \\ h.c.vanhoof@uva.nl 
\And Efstratios Gavves \\ University of Amsterdam \\ egavves@uva.nl
\And Christoph Salge \\ University of Hertfordshire \\ c.salge@herts.ac.uk 
}

\newacronym{RL}{RL}{Reinforcement learning}
\newacronym{MARL}{MARL}{Multi-agent reinforcement learning}
\newacronym{SP}{SP}{Self-play}
\newacronym{IM}{IM}{Intrinsic motivation}
\newacronym{TE}{TE}{Transfer Empowerment}
\newacronym{SI}{SI}{Social influence}
\usepackage[switch]{lineno} 
\begin{document}

\maketitle

\begin{abstract}
We consider multi-agent reinforcement learning (MARL) for cooperative communication and coordination tasks. 
MARL agents can be brittle because they can overfit their training partners' policies. This overfitting can produce agents that adopt policies that act under the expectation that other agents will act in a certain way rather than react to their actions.
Our objective is to bias the learning process towards finding reactive strategies towards other agents' behaviors. 
Our method, transfer empowerment, measures the potential influence between agents' actions.
Results from three simulated cooperation scenarios support our hypothesis that transfer empowerment improves MARL performance. We discuss how transfer empowerment could be a useful principle to guide multi-agent coordination by ensuring reactiveness to one's partner.
\end{abstract}
 
\input{introduction}

\input{related_work}

\input{preliminaries}
\input{methodology}

\input{experiments}

\input{conclusion}
\bibliographystyle{apalike}
\bibliography{bib}
\input{appendix}
\end{document}

%% file: introduction.tex
\section{Introduction}\label{se: intro}

In this paper we investigate if and how social intrinsic motivation can improve \ac{MARL}. \ac{MARL} holds considerable promise to help address a variety of cooperative multi-agent problems - both for problem solving and simulation of multi-agent systems. However, one problem with \ac{MARL} is that agents develop strong policies that are overfitted to their partners' behaviors. Specifically, with centralised training, agents can adopt strategies that expect other agents to act in a certain way rather than reacting to their actions. Such systems are undesirable as they may fail when their partners alter their strategies or have to collaborate with novel partners, either during the learning or deployment phase. Our aim is to avoid this specific lack of robustness and find a guiding principle that makes agents stay reactive to other agents' policy changes. 

We want to introduce an additional reward to bias learning towards socially reactive strategies which should fulfil the following constraints: 1) it should, with minimal adaptation, apply to a wide range of problems with different sensor-actuator configurations to preserve the universality of the RL framework, and 2) it should not negatively affect the performance, i.e., once good policies are found, it should not harm exploitation. Fulfilling the above criteria would provide a general-purpose multi-agent learning algorithm for various cooperative tasks - as well as provide insights into general principles that would enable and improve the development of various forms of social interaction. 

To address this challenge, we turn towards the idea of using \ac{IM} - a school of computational models \citep{oudeyer2009intrinsic} that try to capture the essential motivations behind the behavior of (biological) agents - and then use them for behavior generation to obtain plausible and beneficial behavior. The core idea here is to ask if the principles that create single agent behavior can also be used to enhance multi-agent behavior. In this paper specifically, we look at Empowerment, an \ac{IM} that captures how much an agent is able to affect the world it can itself perceive. Its information-theoretic formulation as the channel capacity between an agent's actions and its own sensors makes it a versatile measure that can be applied to a wide range of models where agent's are defined - satisfying constraint 1. Existing work on coupled empowerment maximisation \citep{salge2017empowerment,guckelsberger2018new} extends the formalism to a multi-agent setting. In this paper, we focus specifically on the idea of \ac{TE}, which tries to capture how much one agent can potentially influence the actions of another.

Keeping the \ac{TE} high between two agents means they are in a state were one of them is reliably reacting to the other. Adding this as an additional reward mechanism during training should help to avoid the brittleness of over-fitting we outlined before. We provide here quantitative evidence for our hypothesis that adding transfer empowerment as an additional reward increases upon the performance of state-of-the art \ac{MARL} methods. Constraint 2 will be evaluated empirically. We also compare this approach to a similar idea of social influence by \citet{jaques2019social}. First, we will introduce the concepts of \ac{MARL} and \ac{IM} in more detail. We will then define the specific formalism for \ac{TE} used, and then simulate three increasingly harder, multi-agent, cooperation scenarios. We will also look at how the better reward was obtained, and discuss the difficult switch from a indexical to an action oriented communication strategy.

%% file: related_work.tex
\section{Related work}
\subsection{Multi-Agent Reinforcement Learning}
There is a large body of research on constructing agents that are robust to their partners. In self-play, for example, agents train against themselves rather than a fixed opponent strategy to prevent developing exploitable strategies \citep{tesauro1994td}. 
Population based-training goes one step further by training agents to play against a population of other agents rather than only a copy of itself. For instance, some methods train an ensemble of policies with a variety of collaborators and competitors \citep{max2018human, lowe2017multi}. 
By using a whole population rather than only a copy of itself, the agent is forced to deal with a wide variety of potential strategies instead of a single strategy.
However, it requires a great deal of engineering because the policy parameters suitable for the previous environment are not necessarily the next stage's best initialization. 

Some works combine the minimax framework and \ac{MARL} to find policies that are robust to opponents with different strategies. Minimax is a concept in game theory that can be applied to find an approach that minimizes the possible loss in a worst-case scenario \citep{osborne2004introduction}. \citet{li2019robust} use it during training to optimize the reward for each agent under the assumption that all other agents act adversarial. 
We are interested in methods that can deal with perturbations in the training partners' behavior, which differs from dealing with partners with various strategies. 

Recent works look at settings in which one RL agent, that is trained separately, must join a group of new agents \citep{lerer2018learning, tucker2020adversarially, carroll2019utility}.  For example, \citet{carroll2019utility} build a model of the other agents which can be used to learn an approximate best response using RL.
\citet{lupu2021trajectory} propose to generate a large number of diverse strategies and then train agents that can adapt to other agents' strategies quickly using meta-learning. 
A related problem is zero-shot coordination \citep{hu2020other} in which agents need to cooperate with unseen partners at test time.
The focus of our paper is not to perform well with novel partners at test-time or build complex opponent models. Our aim is to train agents together to remain attentive and reactive towards their partners' policies.


\subsection{Intrinsic Social Motivation}
Due to centralized training in MARL, agents might adopt non-reactive strategies that may struggle with other agents' changing behaviors. Social intrinsic motivation can give an additional incentive to find reactive policies towards other agents.

\ac{IM} in \ac{RL} refers to reward functions that allow agents to learn interesting behavior, sometimes in the absence of an environmental reward \citep{chentanez2005intrinsically}. 
Computational models of \ac{IM} are generally separated into two categories 
\citep{baldassarre2013intrinsically}, those that focus on curiosity \citep{burda2018large, pathak2017curiosity} and exploration \citep{gregor2016variational, eysenbach2018diversity}, and those that focus on competence and control \citep{oudeyer2009intrinsic, karl2017unsupervised}. The information-theoretic Empowerment formalism \citep{klyubin2005all} is in the latter category, trying to capture how much an agent is in control of the world it can perceive. Empowerment has produced robust behavior linked to controllability, operationality and self-preservation - in both robots \citep{van2020social, karl2017unsupervised, leu2013corbys} and simulations \citep{guckelsberger2016intrinsically}, with \citep{de2018unified} and without \citep{guckelsberger2018new} reinforcement learning and neural network approximations \citep{karl2017unsupervised}.

Empowerment has also been applied to multi-agent simulations, under the term of coupled empowerment maximization \citep{guckelsberger2016intrinsically}, in which it was used to produce supportive and antagonistic behavior. Of particular interest is the idea of transfer empowerment - a measure that quantifies behaviors such as collaboration, coordination, and lead-taking \citep{salge2017empowerment}. 

Similar techniques quantify the interaction between agents for improving coordination between agents. \citet{barton2018measuring} analyze the degree of dependence between two agents' policies to measure coordination, specifically by using Convergence Cross Mapping (CCM).  
\citet{strouse2018learning} show how agents can share (or hide) intentions by maximizing the mutual information between actions and a categorical goal.  
One notably relevant work is by \citet{jaques2019social} called social influence, which is the influence of one agent on the policies of other agents, measured by the mutual information between action pairs of distinct agents. 
Similarly, \citet{mahajan2019maven}, compute the mutual information between agents’ trajectories and a latent variable that captures the joint behavior. \citet{wang2019influence} compute the mutual information between the transition dynamics of agents.

In contrast to social influence (SI), transfer empowerment considers the \textit{potential} mutual information or channel capacity. When optimizing for \textit{actual} mutual information, its value is bounded from above by the lowest entropy of both agent's action variables. SI might easily interfere with an exploitation strategy and may need regularization once a good strategy is found. On the other hand, empowerment does not have this limitation and the action sets could have very narrow distributions, while still being reactive.


%% file: preliminaries.tex
\section{Model}
First, let us define a general model that captures multi-agent scenarios and lets use define transfer empowerment. Let us consider a Dec-POMDP, an extension of the MDP for multi-agent systems, being both decentralized and partially observable \citep{nair2003taming}. This means that each of the $N$ agents conditions the choice of its action on its partial observation of the world. It is defined by the following tuple: $\langle S, \boldsymbol{A}, T, O, \boldsymbol{O}, R, N \rangle$. $S$ is the set of states and $\boldsymbol{A}=\times_{i \in [1, .., n]} \boldsymbol{A}^i$ the set of joint actions. At each time step, the state transition function $P(s_{t+1}|s_t, \boldsymbol{a}_t)$ maps the joint action and state to a new state. As the game is partially observable, we have a set of joint local observations, $\boldsymbol{O}=\times_{i \in [1, .., n]} \boldsymbol{O}^i$ and an observation function $O$. Each agent $i$ selects an action using their local policy $\pi^i(a_t^i|o^i_t)$. 


We consider fully cooperative tasks, so agents share a reward $r(s_t, \boldsymbol{a_t})$ which conditions on the joint action and state. 
The goal is to maximise the expected discounted return  $J(\boldsymbol{\pi}) = \mathbb{E}_{\tau\sim\pi}\left[R(\tau)\right]=\mathbb{E}_{\tau\sim\pi}\left[\sum_{t=0}^T \gamma^t r_t \right]$, with discount factor $\gamma \in [0,1]$ and horizon $T$. The expectation is taken w.r.t. the joint policy $\boldsymbol{\pi}=[\pi^1,\dots,\pi^N]$ and trajectory $\tau=(\boldsymbol{o}_0, \boldsymbol{a}_0, \dots, \boldsymbol{o}_T)$.

%% file: methodology.tex
\section{Methodology}

This section describes an additional heuristic that biases the learning process in obtaining policies that are reactive to other agents' actions. First, we introduce our specific version of transfer empowerment, which rewards the idea of an agent being responsive to adaptations in the other's policy. Then we explain how to train agents in a multi-agent environment.

\subsection{Transfer Empowerment} \label{se: transfer}
Consider two agents, $j$ and $k$, both taking actions and changing the overall state. Each time agent $k$ acts, the state of agent $j$ is modified, and $ j$'s policy indirectly conditions on $ k$'s actions. The objective of coordination is that by changing the actions of agent $k$, agent, $j$ also \emph{reliably} adapts its actions. Here we look at transfer empowerment, namely the \emph{potential} causal influence that one agent has on another. It is defined for pairs of agents by the channel capacity between one agent's action $a^k_t$ and another agent's action  $a^j_{t+1}$ at subsequent time steps and conditioned on the current state $s_t$, which can be computed by maximizing the mutual information $\mathcal{I}$ between those values, with regards to $\omega^k$ :

\begin{equation}
    \mathcal{E}^{k\rightarrow j}(s_t) = \underset{\omega^k}{\max~} \mathcal{I}
    \left(A^j_{t+1}, A^k_t \middle| s_t \right).
\end{equation}

Here, $\omega^k(a^k_t |s_t)$ is the \emph{hypothetical} policy of agent $k$, that takes an action $a^k_t$ after observing state $s_t$ and influencing $a^{j}_{t+1}$ at a later time step. Note that the policy $\omega^k(a^k_t |s_t)$ that maximises the mutual information is not necessarily used for action generation, but simply to compute the channel capacity by looking at all potential policies for the one with the highest mutual information $\mathcal{I}$ .  

Our version of transfer empowerment differs slightly from the one introduced by \citet{salge2017empowerment}, as we consider the potential information flow, or channel capacity, between two agents' \emph{actions} in subsequent time steps. 
\citet{salge2017empowerment} on the other hand, consider the empowerment between one agent's action and another agent's \emph{sensor} state.
Transfer empowerment to another agent's sensory state captures the direct influence to change the other agent's environment. Using transfer empowerment to another agent's \emph{action}, as we do here, focuses on just the influence that affects the other agent's decision. This influence has to flow through the second agent's sensor and be mediated by their policy. In other words, they have to observe and react to the first agent's actions, which aligns with our goal of biasing a policy towards more reactive strategies.

Transfer empowerment has ties with, but is different from, social influence \citep{jaques2019social}. Social influence is the mutual information between agents' actions. It is high when both action variables have a particular entropy, e.g., policies taking different actions. However,  towards the end of the training, a high entropy policy distribution might be suboptimal. Our method, on the other hand, considers the \textit{potential} and not \textit{actual } information flow, so agents only calculate how they \textit{could} influence and react to each other, rather than carrying out its potential. As such, action sets can have very narrow distributions; as long as the system would still be reactive \textit{if}, those actions change. Therefore it does not interfere with obtaining optimal policies.





\subsection{Multi-Agent Training}

Training with transfer empowerment results in joint policies that are reactive to their partner's actions, because for the value to be high, it requires considering the decisions of others. As such, transfer empowerment rewards a very general idea of coordination that requires paying attention to each other, and reliably reacting to a variation in their actions. While empowerment does not measure how this reaction looks, or even if it is good, combined with the actual reward should lead to the selection of a strategy that both solves the problem while also avoiding the brittleness that comes from not being reactive to the information from other agents' policies. Specifically, we will modify the agents' reward function so that it becomes:

\begin{equation}
\tilde{r}(s_t, \boldsymbol{a_t}) = r(s_t, \boldsymbol{a_t}) + \sum_{s_{t+1}}
P(s_{t+1}|s_t, \boldsymbol{a_t}) ~\sum_{j=1}^N \mathcal{E}^{-j \rightarrow j}(s_{t+1}),
\end{equation} 

where $-j$ means all agents excluding agent $j$. To simplify notation, we will use $j$ instead of $-j\rightarrow j$ in the superscript. The new RL objective becomes:
\begin{align*}
    J(\boldsymbol{\pi}) = \mathbb{E}_{\tau \sim \boldsymbol{\pi}}
        \left[
            \sum_{t=0}^T \gamma^t \tilde{r}(s_t, \boldsymbol{a_t})
        \right].
\end{align*}

This new return motivates the potential influence of information between agents' actions, thereby stimulating them to act informatively and react reliably.


\subsection{Efficient Implementation}
We now introduce an efficient implementation to estimate empowerment. We use $a$ for agent $k$'s action at time $t$ and $a'$ for agent $j$'s action at time $t+1$. Mutual information is defined as:

\begin{align*}
    \mathcal{I}(A,A'|s) 
    &=\text{KL}(p(a,a'|s) \| p(a|s)p(a'|s)) \\
    &=\sum_a \sum_{a'}  p(a,a'|s) \ln \frac{p(a,a'|s)}{p(a|s)p(a'|s)} ,
\end{align*}
where $\text{KL}$ is the KL divergence. We can substitute $p(a,a'|s)$ and cancel out terms:

\begin{align*}
    &\sum_a \sum_{a'}  p(a,a'|s) \ln \frac{p(a,a'|s)}{p(a|s)p(a'|s)} \\
    &= \sum_a \sum_{a'} p(a,a'|s) \ln \frac{p(a|a',s)\cancel{p(a'|s)}}{p(a|s)\cancel{p(a'|s)}} \\
    &= \sum_a \sum_{a'} p(a,a'|s) \ln \frac{p(a|a',s)}{p(a|s)}.
\end{align*}

By choosing a variational approximator $q(a|a',s)$, with the property $\text{KL}(p(a|a',s)||q(a|a',s)) \geq 0$, we obtain a lower bound on the mutual information:

\begin{align*}
    \mathcal{I}(A,A'|s) &\geq
    \hat{\mathcal{I}}(A,A'|s) \\
    &:= \sum_a \sum_{a'} p(a,a'|s) \ln \frac{q(a|a',s)}{p(a|s)} \\
    &= \sum_a \sum_{a'} p(a,a'|s) 
        \left(
            \ln q(a|a',s) - \ln p(a|s)
        \right)\\
    &= \mathbb{E}_{p(a,a'|s)}
        \left[
            \ln q(a|a',s) - \ln p(a|s)
        \right].
\end{align*}

The gradient of the lower bound can be approximated by Monte-Carlo sampling.
Furthermore, the overall training procedure can be implemented efficiently when representing the distributions by neural networks and using gradient ascent. 
So the gradient computed over $S$ samples:

\begin{align*}
    \nabla_\theta \hat{\mathcal{I}}_\theta(A,A'|s) &=
    \nabla_\theta \mathbb{E}_{p(a,a'|s)} 
        \left[
            \ln q_\theta(a|a',s) - \ln \omega_\theta(a|s)
        \right] \\
    &\approx
    \frac{1}{S} \sum_{m=1}^S \nabla_\theta \left(\ln q_\theta(a_m|a_m',s) - \ln \omega_\theta(a_m|s)\right),
\end{align*}
where we substituted $p(a|s)$ with $\omega_\theta(a|s)$ and $q(a|a',s)$ with $q_\theta(a|a',s)$, to denote functions parametrized by $\theta$.

\subsection{Partial Observable}
The objective in the previous section was to estimate the empowerment value for a particular state $s$. However, our main goal is to train policies to be reactive towards the actions of their partners. 
Let a policy for agent $j$ be $\pi^j_\chi$ with parameters $\chi$. As each policy is conditioned on its local observations, the lower bound on mutual information for agent $j$ becomes:

\begin{align}
\mi^j_{\theta, \chi}&(A,A'|\boldsymbol o) = \nonumber\\
        &\mathbb{E}_{\boldsymbol o'\sim p_\nu, a^j\sim\pi^j_{\chi}, a^k\sim\omega^k_{\theta}}
        \left[
            \ln q_\theta(\boldsymbol a|\boldsymbol o, \boldsymbol o', a'^j) -\ln\boldsymbol\omega_\theta^{-j}(\boldsymbol a^{-j}|\boldsymbol o^{-j})
        \right],    
        \label{eq: mu}
\end{align}


where samples are generated by a learned transition model $\boldsymbol o' \sim p_\nu(\boldsymbol o'|\boldsymbol o, \boldsymbol a)$. The actions are selected by the target policy $a^j\sim\pi^j_\chi(a^j|o^j)$ and behavior policy $a^k\sim\omega^k_\theta(a^k|o^k)$ and the joint action is $\boldsymbol a = (a^1,\dots, a^j,\dots,a^N)$ where $a^j\sim\pi_\chi$, $a^k\sim \omega_\theta$  and $k\neq j$.

Notice that the actions come from $\boldsymbol \omega_\theta$ and $\pi_\chi$. The former is the joint behavior policy and the latter is the target policy of agent $j$. The behavioral policy is only used to train agent $j$'s policy with empowerment but will not generate extrinsic environmental rewards.

This training procedure has two interesting properties. 
First, it estimates a state's empowerment value. This is done by increasing the diversity of agents' actions while ensuring that these are retrievable from agent $j$'s actions. Actions that affect $j$'s policy, e.g., informative, are chosen more often than those with a lower effect.
Second, it trains agent $j$'s policy to be reactive towards the actions of its partners, because we compute the gradient of mutual information w.r.t. $\chi$ to directly optimize $\pi_\chi$. We provide the description of the full algorithm in the Appendix, which also describes how our method applies to settings with more than 2 agents.

Altogether, empowerment prefers states that allow for information flow between agents, altering policies to be more responsive. We will experimentally verify this in the next section.

%% file: experiments.tex
\section{Experimental Results}
We adopt the simulator developed for testing multi-agent reinforcement learning algorithms \footnote{https://github.com/openai/multiagent-particle-envs} that allows creating cooperative and competitive environments. Agents have a continuous observation space and a discrete actions space. 

\begin{figure*}[!hb]
    \centering
    \begin{subfigure}[b]{0.3\textwidth}
         \centering
         Simple
         \includegraphics[width=\textwidth]{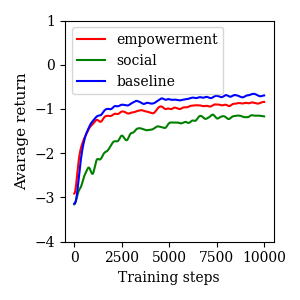}
     \end{subfigure}
     \begin{subfigure}[b]{0.3\textwidth}
         \centering
         Challenging
         \includegraphics[width=\textwidth]{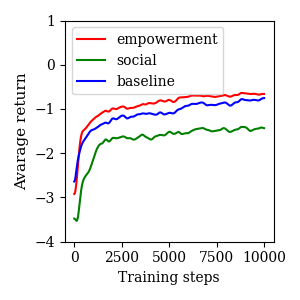}
     \end{subfigure}
     \begin{subfigure}[b]{0.3\textwidth}
         \centering
         Hard
         \includegraphics[width=\textwidth]{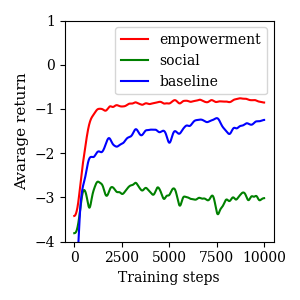}
     \end{subfigure}
        \caption{Learning curves for the the three tasks. The rewards are averaged over the steps in an episode to obtain the return. The returns are averaged over three training runs.}
        \label{fig:loss}
\end{figure*}

\subsection{Scenarios}
\begin{figure}[!h]
    \centering
    \begin{subfigure}[b]{0.15\textwidth}
         \centering
         Simple
         \includegraphics[width=\textwidth]{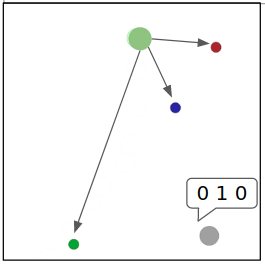}
     \end{subfigure}
     \begin{subfigure}[b]{0.15\textwidth}
         \centering
         Challenging
         \includegraphics[width=\textwidth]{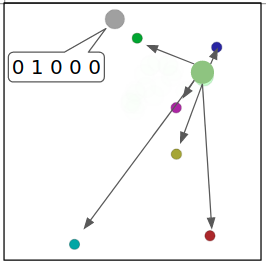}
     \end{subfigure}
     \begin{subfigure}[b]{0.15\textwidth}
         \centering
         Hard
         \includegraphics[width=\textwidth]{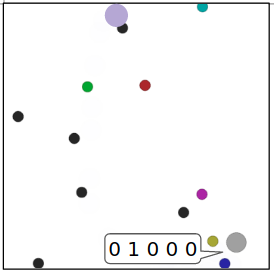}
     \end{subfigure}
        \caption{Visualizations of the three tasks. Small dark circles indicate landmarks and obstacles, and the big circles are the listener and speaker. The listener observes the relative distance to the landmarks indicated by the arrows. The speaker's massages are one-hot vectors displayed by the speaker boxes.}
        \label{fig:tasks}
\end{figure}

We use a cooperative two-dimensional environment consisting of two agents. The (disembodied) speaker agent can, each time step, choose a communication action that is broadcast to the listener. The listener can choose from 5 physical actions, moving up, down, left and right, or doing nothing. The environment contains a series of $L$, randomly placed, landmarks. Only the speaker has a signal informing it which of the $L$ landmarks is the target. The objective for the randomly placed listener is to reach the target landmark by decoding the speaker's message. The speaker can send a symbol chosen from a set of $C$ distinct symbols. The team reward is the negative squared distance between the listener and the target landmark, which is given out every time step. The game ends after 100 time steps. To perform well the listener has to quickly move onto the landmarks. We developed three tasks in the environment with increasing difficulty. Fig. \ref{fig:tasks} visualises the tasks. 

\paragraph{Simple}
The number of symbols $K=|C| = 3$ equals the number of landmarks $L=3$. The speaker observes the color of the target landmark, while the listener sees a distance vector pointing to each colored landmark.

This scenario could be solved well by an indexical communication strategy, where the speaker simply has to consistently assign a symbol to each landmark color, and then simply relay the information to the speaker, who then has to minimise the distance to the landmark of that color. 


\paragraph{Challenging}
The second task involves more landmarks $L=6$ than distinct symbols $K=|C|=5$. The speaker observes the target \emph{position} and the listener's position, while the listener observes the landmarks' positions and the messages sent by the speaker.
Here, an \textit{action-oriented strategy}, e.g., indicating movement direction, is likely optimal because the speaker cannot use each symbol for a landmark uniquely, nor do the landmarks have any identifying features that are easy to community, i.e. they are not colored anymore. Using symbols to direct the listener now requires the speaker to observe and react to the listener's position with an updated signal, and put more cognitive demands on the speaker, who could simply relay its internal signal in the simple scenario. 

 
\paragraph{Hard}
The last task adds $M=6$ obstacles, and the reward includes a penalty if the listener hits an obstacle. Furthermore, the landmarks' positions are now unobserved by the listener. These two features increase the difficulty because a higher precision is required. First, the listener has to avoid obstacles, and second, the listener is even more dependent on the speaker because it does not see the landmarks.

\paragraph{Reward}
The reward function is determined from the position, $\pos=[p_x,p_y]$, of the listener (agent 1) $\pos^1$, target $\pos^g$ and obstacles $\pos^o$. The state is defined as $s_t = [\pos_t^1, \boldsymbol{m}_t, \pos^g, \pos^{o, 1}, \dots, \pos^{o, M}, \pos^{l, 1}, \dots, \pos^{l, L}]$ for $M$ obstacles, and $L$ landmarks. The reward function is:
\begin{align}
    r(s_t, \boldsymbol{a}_t) &= -\|\pos^1_{t+1} - \pos^g\| + \text{penalty} \\
     \text{penalty} &= 
     \begin{cases}
        -1 & ~ \text{if } \quad \exists j \in [1,...,M]: \|\pos^1_{t+1} - \pos^{o,j}\| <0.15\\
        0  & ~ \text{otherwise}
    \end{cases}
\end{align}
The observation for the speaker and listener $o_t^0 = [\pos^1_t, \pos^g, \pos^{l, 1}, \dots, \pos^{l, L}]$ and $o_t^1 = [\boldsymbol{v}^1_t, \boldsymbol{m}^0_t]$, respectively. $v$ is the velocity and $m$ the message, represented by a one-hot vector. 
The listener's action is a force vector $a^1=[f_x, f_y]$, while the speaker's action is a message $a^0=[c^0, \dots, c^{K}]$ with vocabulary size $K$.
The listener's position is updated according to the following equation:
\begin{align}
        \begin{bmatrix}
            \boldsymbol{p} \\
            \boldsymbol{v} \\
            \boldsymbol{\dot{v}}
        \end{bmatrix}_{t+1}
        = 
        \begin{bmatrix}
            \boldsymbol{p} + \boldsymbol{v} \Delta t\\
            \zeta \boldsymbol{v} + \boldsymbol{\dot{v}} \Delta t \\
            \frac{\boldsymbol{u}}{\text{mass}}
        \end{bmatrix}_t
\end{align}
with damping coefficient $\zeta=0.5$ and $\text{mass}=1$.
The speaker's messages, will be added to the state at the next time-step: $\boldsymbol{m}_t = \boldsymbol{a}^0_{t-1}$.
As is common when working with policies parameterised by neural networks, the actions are one-hot vectors, obtained by Gumbel-Softmax function \citep{pml1Book}. For example, the actions of the speaker are converted into
\begin{align}
    \text{one-hot}(\boldsymbol{a^0}) = [\mathbb{I}(a^0_1=\max(\boldsymbol{a^0})),\dots, \mathbb{I}(a^0_K=\max(\boldsymbol{a^0}))].\nonumber
\end{align}

\begin{table}[!hb]
\centering
\caption{The values show the average distance between the listener and target landmark and the percentage of collisions with obstacles. The results are computed for 100 episodes after training with 10k episodes.}
\begin{tabular}{lccccc}
\toprule
    &  \multicolumn{1}{c}{Simple} & \multicolumn{1}{c}{Challenging}&\multicolumn{2}{c}{Hard}\\
    \midrule
{}              & {Average}  & {Average}          & {Average} & {Obstacle} \\
{}              & {distance} &   {distance}         & {distance} & {hit } &\\
\midrule
{basel.}      & 0.221     &  0.440                   & 0.520        & 0.603     \\
{SI}  & 0.716     &  0.949                     & 1.076        & 0.220     \\
{\ac{TE}}   & 0.414     &  0.460                    & 0.440        & 0.266     \\
\bottomrule
\end{tabular}
\label{tab:comm}
\end{table}

\subsection{Comparison and Implementation details}
We compare our method with MADDPG \citep{lowe2017multi} (baseline) and social influence \citep{jaques2019social} (social infl.). Our method (empowerment) is built on top of the MADDPG, a centralized actor-critic method. Social influence is a decentralized method.  
The agents' policies are parameterized by a two-layer ReLU MLP with 64 units per layer. The messages sent between agents are soft approximations to discrete messages, calculated using the Gumbel Softmax estimator. 
All models are trained for 10k episodes, of which an episode consists of 25 interactions. Source code can be found at \url{https://github.com/tessavdheiden/social_empowerment}.

\section{Results and Discussion}
\subsection{Learning Curves}
Our two main hypothesis are that adding \ac{TE} to \ac{MARL} produces faster adaptation (needs less training steps),  and achieves better, overall results. To answer both of the questions, we compare the learning curves, over 10k training steps, averaged over three runs, for both the \ac{MARL} baseline, and with the addition of \ac{TE} (empowerment) and \ac{SI} (social). 
Figure \ref{fig:loss} shows the averaged return after a given number of training steps. A higher score is better, it shows that the listener is closer to the target. Since the listener starts away from the target a score of 0 is impossible, all scores are  negative. 

The learning speed seems to be comparable between models in difference scenarios, i.e. it takes about the same time for the different algorithm to reach their peak, final performance. Only \ac{SI} seems to learn slower in both the simple and challenging task.
Performances seem to mostly stabilise after some point, so we can also take a closer look at the performances of the trained agents after 10k training steps.

\begin{figure}
     \centering
     \begin{subfigure}[b]{0.23\textwidth}
         \centering
         Baseline
         \includegraphics[width=\textwidth]{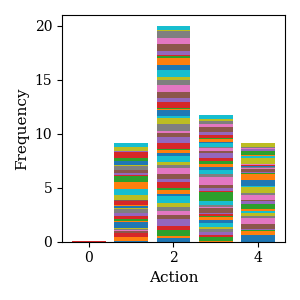}
     \end{subfigure}
     \begin{subfigure}[b]{0.23\textwidth}
         \centering
         Empowerment
         \includegraphics[width=\textwidth]{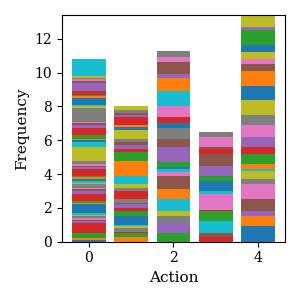}
     \end{subfigure}
        \caption{The action distributions for the listener in the hard task for both the baseline and empowerment. The colors indicate each a different episode. 0 is the wait action, 1 - 4 are cardinal accelerations.}
        \label{fig:actions}
\end{figure}

\subsection{Final Scores}
Details of the results are presented in Table \ref{tab:comm}. It shows the average distance and the percentage of collisions with an obstacle for the final agents, computed for 100 episodes. 

The baseline obtains the top performance, with lowest distance of $0.221$, in the simple task, requiring a simple, lexical strategy. Here empowerment performs worse with $0.414$, with social influence performing even worse with $0.716$. This is an indication of the idea that an added incentive might get in the way of exploiting an easy to find, simple strategy.

In the challenging task, which requires an action-oriented strategy empowerment seems to perform similar to the baseline, while it clearly outperforms the baseline for the hard task, both in terms of average distance, and in terms of obstacles hit. The difference in hit obstacles is particularly large, indicating that \ac{TE} helps with the higher reactivity required within an episode to navigate around the obstacles. Social influence seems to struggle with both tasks, again likely due to an interference between the added reward and the best exploitation strategy. 

In contrast, the better performance of empowerment in the hard challenge, compared to the baseline, must be due to empowerment helping to discover a better overall strategy - as the baseline implementation would be fully capable of producing a strategy identical to the one performance by the \ac{TE} framework, had it discovered it.  

To illuminate this difference, we can take a look at the action distribution for the speaker and listener agents over several episodes, using the agents after 10k training episodes. Fig.~\ref{fig:actions} shows how often the five available actions were used. Action 0 for the listener is the waiting action - and we see that this one is not used by the baseline. We speculate that it might be difficult for the listener to learn when to use this action, as it is detrimental in most cases. The bias towards reactivity induced by \ac{TE} might help to keep this rare action as an option - following a symbol by the speaker that might become a ``stop'' signal. 

\begin{figure}
     \centering
     \begin{subfigure}[b]{0.15\textwidth}
        \centering
        Baseline
        \includegraphics[width=\textwidth]{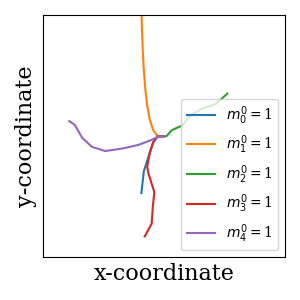}
     \end{subfigure}
     \hfill
     \begin{subfigure}[b]{0.15\textwidth}
         \centering
         Empowerment
         \includegraphics[width=\textwidth]{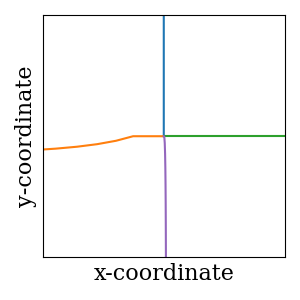}
     \end{subfigure}
     \hfill
     \begin{subfigure}[b]{0.15\textwidth}
        \centering
        Social influence
        \includegraphics[width=\textwidth]{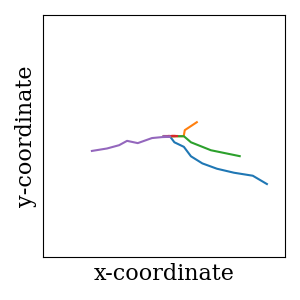}
     \end{subfigure}
        \caption{The listener's positions plotted for 10 time steps, given a speaker's message $\boldsymbol{m}^0$, a one-hot vector. The subscript denotes the component in $\boldsymbol{m}^0$  that is equal to a 1.}
        \label{fig:positions}
\end{figure}

We can also take three trained listener agents and compare what they will do when we provide them with a fixed speaker signal over several time steps, to figure out what those symbols directed them to do. Fig.~\ref{fig:positions} shows the trajectories resulting from this, with each color denoting a different forces symbol by the speaker. The baseline has a relatively good separation into cardinal actions, but transfer empowerment leads to nearly perfect control by the speaker over the actions of the listener. Note that one signal results in the wait action, leading to no visible trajectory for empowerment.

%% file: conclusion.tex
\section{Conclusions and Future Work}

Overall, adding transfer empowerment to \ac{MARL} seems to improve the overall performance level of cooperative agents - particularly for harder tasks that rely on an action-oriented communication strategy. This seems to indicate that \ac{TE} helps the learning process to find better solutions to converge on - which remain undiscovered by the baseline \ac{MARL} with similar training time - while also not getting in the way of exploitation to much. An immediate open question, a direction for future work, is of course the question of generalizability of this approach to different scenarios. Other exciting research directions are scenarios with partners unseen at training time, moving in the direction of one-shot adaptation to partners, and scenarios with competitive, or cooperative-competitive mixed scenarios. Using \ac{TE} to bias systems towards control, or information hiding to find optimal solutions. 

We also showed how an efficient computation of empowerment could be combined with \ac{RL} for the \ac{MARL} framework, opening the door for more complex scenarios such as humans interacting with robots. In general, the results in this study are promising for the overall agenda to develop a framework of social intrinsic motivations based on empowerment (or similar measures) to bias an agent towards general social concepts, such as reliable reactivity, or lead-following. The fact that it is based on similar, single-agent intrinsic motivations is also interesting, as it might offer insights into how to transition from single to social agent behavior with only gradual adaptation.

%% file: appendix.tex
\newpage
\onecolumn
\section*{Appendix}
\label{se: appendix}

Algorithm \ref{alg: train} explains how we train with empowerment. We omit super- and subscripts denoting time, agent and batch indices whenever clear from the context. 

\newcommand{\var}{\texttt}

\begin{algorithm*}
    \caption{Training joint policy $\boldsymbol \pi_\theta$ with empowerment}\label{alg:training}
    \begin{algorithmic}
    \Require Initialisation of networks $\boldsymbol \pi_\chi$, $\boldsymbol Q_\psi$, $\boldsymbol \omega_\theta$, $q_\theta$ and $p_\nu$, and target networks $\boldsymbol\pi_\phi$, $\boldsymbol Q_\zeta$.
    \ForEach{episode}
        \ForEach{time step}
            \State $\boldsymbol o = O(s)$, $\boldsymbol a \sim \boldsymbol \pi(\boldsymbol a|\boldsymbol o)$, $s'\sim f(s, \boldsymbol a)$
            \State $\tau = \tau \cup \{(\boldsymbol o, \boldsymbol a, \boldsymbol o', r, \mi, y)\}$
        \EndFor
        \State $\mathcal{D} = \mathcal{D}  \cup \tau $
        \ForEach{agent $i$}
                \State sample minibatch with $S$ tuples from $\mathcal{D}$
                \State $\mi_{\chi^i, \theta^i}(\boldsymbol o)=$ \var{computeLowerBound}($\boldsymbol o$,$\boldsymbol \pi_\chi$, $\boldsymbol \omega_\theta$, $q_\theta$, $p_\nu$) \Comment{Equation \ref{eq: mu}}
            \State $y = r + \mi_{\chi^i, \theta^i}(\boldsymbol o) + \gamma Q_\zeta^i(\boldsymbol o, \boldsymbol a)$
                \State $\mathcal{L}(\psi^i) = \frac{1}{S}\sum_j 
                    (
                        y - Q^i_\psi(\boldsymbol o_j, \boldsymbol a_j)
                    )^2|_{a^i_j \sim \pi^i_\phi}$
                \State $\mathcal{L}(\chi^i) = -\frac{1}{S}\sum_j 
                    Q_\psi^i(\boldsymbol o_j, \boldsymbol a_j)|_{a^i_j \sim \pi^i_\chi}$

                \State \var{updateCritic}($\mathcal{L}(\psi^i)$, $\psi^i$) \Comment{See \citep{lowe2017multi}}
                \State \var{updateActor}($\mathcal{L}(\chi^i)$, $\chi^i$)
                \State \var{gradientAscent}($\mi_{\chi^i, \theta^i}$, $\theta^i$, $\chi^i$)
                \State \var{maxLogLikelihood}($\nu$, $\boldsymbol o$, $\boldsymbol a$, $\boldsymbol o'$)  \Comment{See \citep{karl2016deep}}
        \State \var{updateTargets}($\phi^i$, $\zeta^i$, $\psi^i$, $\chi^i$)
        \EndFor
    \EndFor
    \end{algorithmic}
    \label{alg: train}
\end{algorithm*}

\begin{table*}[!hb]
    \centering
    \begin{tabular}{l|l}
    \hline
        $\boldsymbol{\pi}$              & Joint policy with $N$ components $[\pi^1, \dots, \pi^N]$. \\
        $O(s)$            & Deterministic observation function $\boldsymbol o = (o^1,\dots,o^N) = O(s)$.\\
        $\boldsymbol a, \boldsymbol o$  & Joint action and observation $(a^1, \dots, a^N)$, $(o^1,\dots,o^N)$.\\
        $\boldsymbol a^{-i}, \boldsymbol o^{-i}$            & Joint action and observation excluding those of agent $i$.\\
        $J^i(\boldsymbol{\pi})$         & Expected return of agent $i$ induced by joint policy $\boldsymbol{\pi}$. \\
        $Q^i(\boldsymbol{o}, \boldsymbol{a})$      & Centralised critic of local policy $\pi^i$. \\
        $\mathcal{E}^{-i\rightarrow i}$ & Transfer empowerment from all agents' actions, excluding agent $i$, towards agent $i$'s action.\\
        $\hat{\mathcal{E}}$ & Lower bound on empowerment by employing variational approximation. \\
        $\hat{\mathcal{I}}_\theta$ & Lower bound on mutual information computed by $\theta$-parameterized neural networks.
    \\ \hline
    \end{tabular}
    \caption{Notation}
    \label{tab:notation}
\end{table*}